%% file: aaai25.tex
\documentclass[letterpaper]{article} % DO NOT CHANGE THIS
\usepackage{aaai25}  % DO NOT CHANGE THIS
\usepackage{times}  % DO NOT CHANGE THIS
\usepackage{helvet}  % DO NOT CHANGE THIS
\usepackage{courier}  % DO NOT CHANGE THIS
\usepackage[hyphens]{url}  % DO NOT CHANGE THIS
\usepackage{graphicx} % DO NOT CHANGE THIS
\usepackage{natbib}  % DO NOT CHANGE THIS AND DO NOT ADD ANY OPTIONS TO IT
\usepackage{caption} % DO NOT CHANGE THIS AND DO NOT ADD ANY OPTIONS TO IT
\usepackage{algorithm}
\usepackage{algorithmic}

\usepackage{xcolor}
\usepackage{colortbl}
\definecolor{weblink}{HTML}{F60286}
\newcommand{\trans}{symmetric power transformation}
\newcommand{\Trans}{Symmetric Power Transformation}\usepackage{amsfonts}
\usepackage{amsfonts}
\usepackage{newfloat}
\usepackage{listings}
\usepackage{booktabs}
\usepackage{amsmath}
\DeclareCaptionStyle{ruled}{labelfont=normalfont,labelsep=colon,strut=off} % DO NOT CHANGE THIS
\floatstyle{ruled}
\newfloat{listing}{tb}{lst}{}
\floatname{listing}{Listing}
\title{Enhancing Implicit Neural Representations via Symmetric Power Transformation}
\author{
    Weixiang Zhang,
    Shuzhao Xie,
    Chengwei Ren,
    Shijia Ge,
    Mingzi Wang,
    Zhi Wang\textnormal{\textsuperscript{*}}
}
\affiliations{
    Tsinghua Shenzhen International Graduate School, Tsinghua University, Shenzhen, China \\ 
    \{zhang-wx22, xsz24, rcw22, gsj23, wmz22\}@mails.tsinghua.edu.cn, wangzhi@sz.tsinghua.edu.cn \\
    \textcolor{blue}{\texttt{https://weixiang-zhang.github.io/proj-symtrans}}
    
}

\usepackage{bibentry}
\begin{document}

\maketitle
\input{sections/0_abstract}
\input{sections/1_intro}

\input{sections/3_method}

\input{sections/4_experiment}
\input{sections/2_related}
\input{sections/5_conclusion}

\clearpage
\section*{Acknowledgments}
This work is supported in part by National Key Research and Development
Project of China~(Grant No. 2023YFF0905502), National Natural Science Foundation of China~(Grant No. 62472249), and Shenzhen Science and Technology Program~(Grant No. JCYJ20220818101014030).
We thank anonymous reviewers for their valuable advice.

\bibliography{aaai25}

\end{document}

%% file: sections/0_abstract.tex
\begin{abstract}
We propose \trans \ to enhance the capacity of Implicit Neural Representation~(INR) from the perspective of data transformation. 
Unlike prior work utilizing random permutation or index rearrangement, our method features a reversible operation that does not require additional storage consumption.
Specifically, we first investigate the characteristics of data that can benefit the training of INR, proposing the Range-Defined Symmetric Hypothesis, which posits that specific range and symmetry can improve the expressive ability of INR. 
%This hypothesis is verified using both simulated and real data.
Based on this hypothesis, we propose a nonlinear \trans~to achieve both range-defined and symmetric properties simultaneously. We use the power coefficient to redistribute data to approximate symmetry within the target range. To improve the robustness of the transformation, we further design deviation-aware calibration and adaptive soft boundary to address issues of extreme deviation boosting and continuity breaking.
Extensive experiments are conducted to verify the performance of the proposed method, demonstrating that our transformation can reliably improve INR compared with other data transformations. We also conduct 1D audio, 2D image and 3D video fitting tasks to demonstrate the effectiveness and applicability of our method.
Code is available at \textcolor{blue}{https://github.com/zwx-open/Symmetric-Power-Transformation-INR}.

\end{abstract}
% \begin{links}
% \link{Code}{https://github.com/zwx-open/Symmetric-Power-Transformation-INR}
% \end{links}

%% file: sections/1_intro.tex
\section{Introduction}
Implicit Neural Representations~(INRs) have been proposed for continuously representing signals using neural networks, which has garnered significant attention in the representation of multimedia data, including audios~\cite{inras}, images~\cite{img-inr} and videos~\cite{nerv}. Compared to traditional discrete representations such as amplitudes, pixels and frames, INR offers several advantages, including memory efficiency, resolution agnosticism, and suitability for inverse problems. These advantages make INR particularly useful in downstream tasks such as compression~\cite{inr-compress}, super-resolution~\cite{event-super-inr}, novel view synthesis~\cite{nerf}, and physics simulations~\cite{pinn}.

\begin{figure}[!t]
    \centerline{\includegraphics[width=\columnwidth]{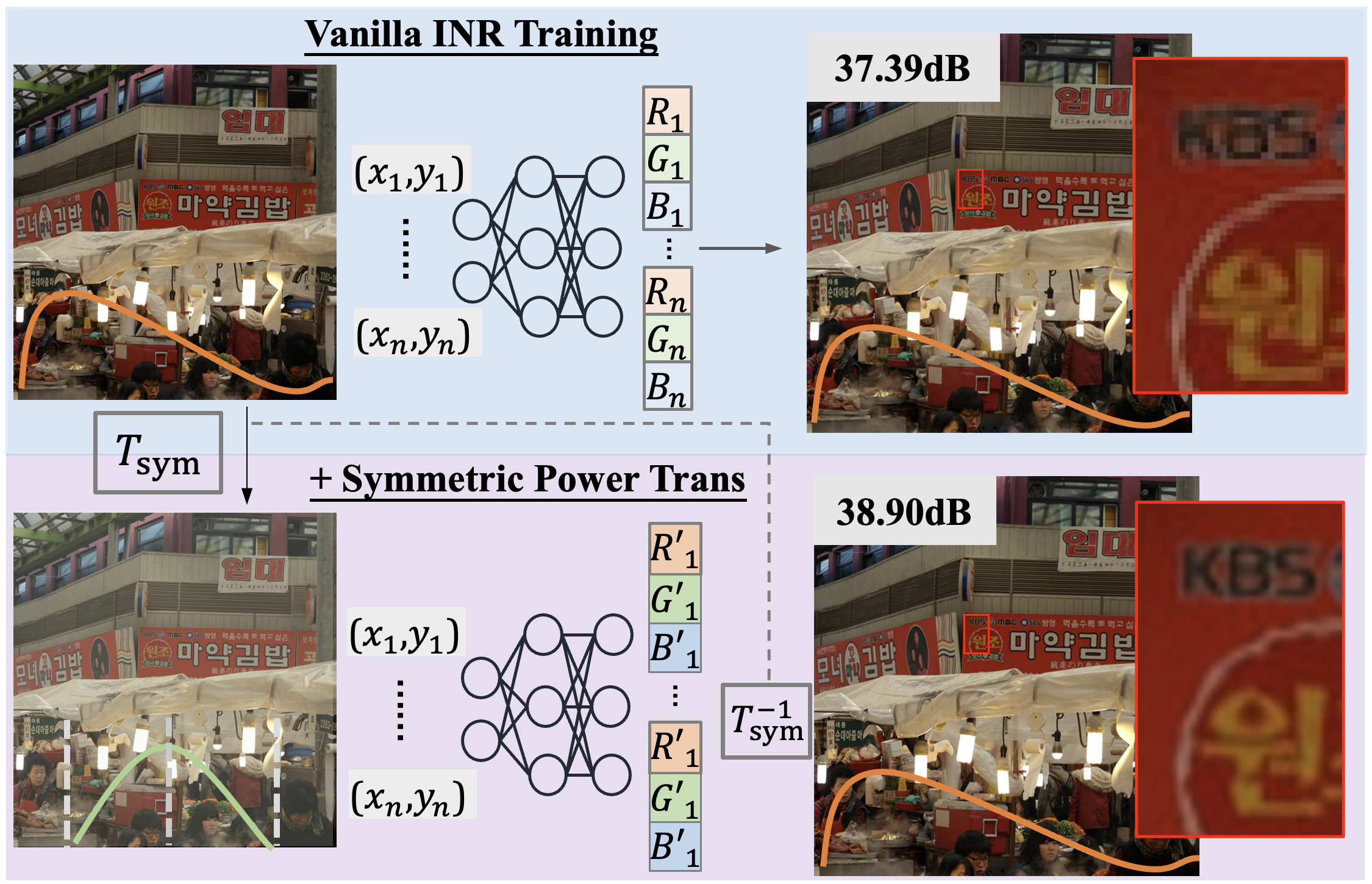}}
    \caption{Pipeline of \Trans. We transform data with specific range and symmetry to  enhance implicit neural representation without any additional time and spatial consumption. Integrating this simple nonlinear transformation can significantly improve the reconstruction quality. \textit{See comparison of ``KBS'' in reconstructed image.}}
    \label{fig:intro}
\end{figure}

However, encoding signals into neural representations is resource-intensive, requiring training a neural network to fit natural signals, which hinders the application of this novel representation in real-world scenarios. 
Therefore, optimizing the training process of INR has become a crucial research topic  in recent times. 
To address this challenge, various approaches have been proposed to improve INR training from different perspectives, including periodic activation functions~\cite{finer}, partition-based acceleration~\cite{miner}, and reparameterized training~\cite{reparam}.
Recently, RPP~\cite{search} provided an insight to enhance INR through \textit{data transformation} in the pre-processing stage, aiming to transform data such that optimization more effectively fits the transformed data with INR.
However, this work primarily focused on random pixel permutation~(RPP), which, although accelerating training in the long run, disrupts signal continuity and thus lowers performance in typical training scenarios, limiting real-world applicability. 
Moreover, such index rearrangement necessitates an additional mapping to store the projections of the index pair before and after the pixel permutation. This requirement results in significant storage consumption comparable to the input signal, thereby diminishing the memory efficiency advantages.

To overcome these limitations, we propose \trans, a cost-free data transformation to enhance implicit neural representations. 
We first investigate the characteristics of data that can benefit the training of INR. Based on intuition of periodic activation functions, we review the data from the range and symmetry perspectives, proposing the Range-Defined Symmetric Hypothesis. Specifically, scaling data to within a specific range and redistributing the data symmetrically can enhance the expressive capability of INR. 
Based on this hypothesis, we design \trans~to improve the training of INR. We select power transformation~($x \to x^\beta$) as the basic form due to its reversibility and monotonicity. By normalizing data to $[0,1]$, we ensure the transformed values remain bounded. The power coefficient $\beta$ serves to redistribute data and approximate symmetry in the distribution. Given the complexity of natural signals, we additionally propose two methods to enhance the robustness of \trans: deviation-aware calibration and adaptive soft boundary. These methods respectively mitigate extreme deviation boosting and continuity-breaking at boundaries, ensuring consistent performance improvements.

Extensive experiments were conducted to verify the performance of our method. 
Our method achieved the most significant improvement compared to all other current transformations, and its integration consistently enhances performance in 1D audio, 2D image, and 3D video fitting tasks.
Beyond its capability for steady improvement, the reversibility of \trans~enables lossless signal reconstruction without additional storage consumption, making it a \textbf{cost-free} enhancement for INR. This is the most significant advantage compared to prior work.

The main contributions are summarized as follows:
\begin{itemize}
    \item We observe that scaling the data to a specific range and ensuring a symmetric distribution benefits the training of INRs. Based on these observations, we propose the Range-Defined Symmetric Hypothesis.
    \item We propose \trans~to enhance implicit neural representation. We also introduce deviation-aware calibration and adaptive soft boundary to further improve the robustness of the method.
    \item We verify the effectiveness of our method through extensive experiments, including 1D audio fitting, 2D image fitting, and 3D video fitting task.
\end{itemize}

%% file: sections/3_method.tex
\section{Method}
\subsection{Background and Problem Formulation}
Consider an Implicit Neural Representation~(INR) function: $F_{\theta}: \mathcal{X} \in \mathbb{R}^c \mapsto \mathcal{Y} \in \mathbb{R}^{s}$.
This function maps a $c$-dimension coordinate $\mathcal{X}$ to the discrete form of a $s$-dimension signal $\mathcal{Y}$, with $\theta$ representing the optimizable parameters a neural network.
Given a signal set $\mathcal{S} = \left\{\mathbf{{x}_i}, \mathbf{{y}_i}\right\}_{i=1}^N$ composed of $N$ pairs of coordinates $\mathbf{{x}_i}$ and their corresponding signal representations $\mathbf{{y}_i}$, the goal is to overfit an $L$-layer MLP $F_{\theta}(\mathbf{x})$ to the ground truth $\mathbf{y}$ with minimal loss. 
The output of layer $l$ is given by $\mathbf{z}_l=\sigma\left(\mathbf{W}_l \mathbf{z}_{l-1}+\mathbf{b}_l\right)$, where $\sigma$ is a nonlinear activation function, and $\mathbf{W}_l$ and $\mathbf{b}_l$ are the weights and bias of layer $l$.
Optimizing $F_{\theta}$ using conventional ReLU activations often fails to achieve satisfactory performance due to Spectral Bias~\cite{bias}, which refers to the tendency of MLPs to fit the low-frequency components of signals more readily than the high-frequency components.
SIREN~\cite{siren} effectively mitigates this issue by employing periodic sine as active functions to represent complex natural signals. The output of the $l$-th layer can be expressed as $\mathbf{z}_l = \sin(\omega_0 (\mathbf{W}_l \mathbf{z}_{l-1} + \mathbf{b}_l))$, where $\omega_0$ controls the frequency of the network. 
To further address frequency issues, FINER~\cite{finer} introduced variable-periodic activation function $sin(( | x | + 1)x)$, achieving state-of-the-art performance.

Given the high capacity for signal representation, we consider SIREN and its variant FINER as the backbone.
Our objective is to enhance its training from the perspective of signal transformation without incurring additional time or space consumption. Mathematically, we aim to identify a reversible transformation $T: \mathcal{Y} \in \mathbb{R}^s \mapsto \mathcal{H} \in \mathbb{R}^s$, with its inverse denoted as $T^{-1}$. Given the transformed signal set $\mathcal{S}^T = \left\{ \mathbf{x}_i, T(\mathbf{y}_i)\right\}_{i=1}^N$, the INR function $F_\phi$ is optimized with the ground truth $T(\mathbf{y})$. Under the same iteration $t$, the transformed data should exhibit improved expressiveness compared to the vanilla data, adhering to the condition $Q(F^t_\theta(\mathbf{x}), \mathbf{y}) < Q(T^{-1}(F^t_\phi(\mathbf{x})), \mathbf{y})$, where $Q(\cdot)$ is the metric used to measure the quality of the reconstruction.

\subsection{Range Defined Symmetric Hypothesis}

In order to determine the target transformation $T$ to enhance implicit neural representations, it is essential to investigate the characteristics of distribution that are beneficial for the training process. We provide insights from the perspectives of range and symmetry.

\begin{figure*}[!t]
    \centerline{\includegraphics[width=\textwidth]{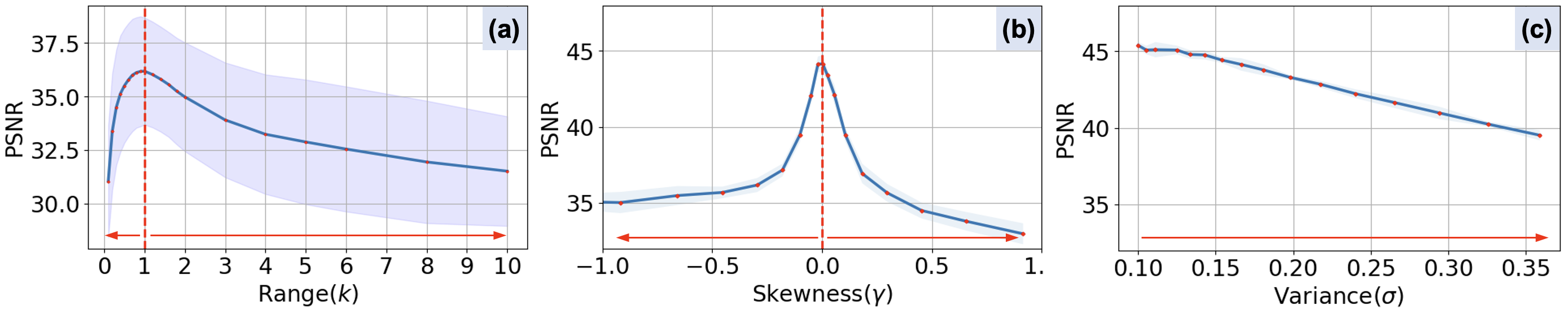}}
    \caption{
    \textbf{(a)\&(b):}  Hypothesis verification from range and skewness perspectives; \textbf{(c):} Investigation of variance effects under zero-skewness distribution.
    Red arrows indicate performance degradation direction.}
    \label{fig:sym-1.1}
\end{figure*}

\textbf{Range.} The output of each layer in INRs is bounded by the range $I$, which confines the final output of the network approximately within this bound. Specifically, the output in the $(L-1)$-th layer $\mathbf{z}_{L-1} \sim I$, results in the final output $F_\theta(\mathbf{x})=\mathbf{W}_L \mathbf{z}_{L-1} + \mathbf{b}_L \sim kI+b$. The initialization scheme~\cite{kaiming-init} in the last layer ensures that $kI+b$ approximates $I$. Therefore, transforming the signal $\mathbf{y}$ within the bound $I$ should be beneficial for INRs training.

\textbf{Symmetry.} Unlike traditional ReLU-based MLPs, activation functions $\sigma(\cdot)$ in INRs are centrally symmetric, satisfying $\sigma(x) = -\sigma(-x)$. Additionally, the initial scheme for weights and biases is symmetric about zero, leading the entire neural network to maintain symmetry during optimization. If we transform the distribution to lower skewness~\cite{skew}, thereby approximating symmetry, the network's performance could be enhanced.

Based on the above intuitions, we propose the Range-Defined Symmetric Hypothesis as follows:

\textbf{Range-Defined Symmetric Hypothesis.} \textit{Given an INR $F_\theta$ with a bounded periodic activation function $\sigma(\cdot) \sim I$ and an input signal $\mathbf{y}$ with distribution $\mathcal{G}$, satisfying the following conditions can enhance the expressive ability of INRs: 
1)~Range-Defined: the bound of $\mathbf{y}$ is approximately $I$. 
2)~Symmetric: the skewness of $\mathcal{G}$ is approximately 0.}

We verify this hypothesis in the context of training INRs from the following perspectives: 

\underline{1.~Different Range.} Given a natural signal $\mathbf{y}$ and a bounded activation function $\sigma\sim I$, we implement min-max normalization to constrain $\mathbf{y}$ within $I$. We then use a coefficient $k$ to generate different bounds $kI$. This verification is conducted using real image datasets~\cite{pe}.

\underline{2.~Different Skewness.} To precisely measure the effect of skewness, we use simulated data to avoid confounding factors. We generate signals $\mathbf{y_N}$, whose intensities are sampled from a normal distribution $\mathcal{N}(\mu, \sigma)$, where $\mu$ and $\sigma$ represent the mean and standard deviation, respectively. 
We scale the data to the range $[-1, 1]$ to satisfy the optimal range settings and keep $\sigma$ constant to exclude frequency effects. By varying $\mu$, we simulate cases with different skewness $\gamma$, which is computed using the third standardized moment. When $\mu = 0$, the distribution achieves zero-skewness. As $|\mu|$ increases, $\gamma$ increases, resulting in greater asymmetry.

\underline{3.~Different Deviation under Zero-Skewness.} We further investigate the impact of variance under symmetric distributions by maintaining zero-skewness and varying $\sigma$ while preserving the settings for $\mathbf{y_N}$.

As Fig. \ref{fig:sym-1.1} (a)\&(b) demonstrate, our hypothesis regarding range and skewness can be verified. The expressive ability of the INR reaches its maximum when $k=1$ and $\gamma=0$, corresponding to the defined range and symmetric status of signals. The performance gradually declines as the distribution deviates from these two configurations.
Based on the exploration of different variances under zero-skewness~(Fig. \ref{fig:sym-1.1} (c)), we further propose a supplementary statement for Range-Defined Symmetric Hypothesis: 

\textbf{Supplementary Statement.} \textit{When setting a symmetric data distribution, the performance declines approximate linearly as deviation increases.} 

This phenomenon can be explained by the fact that an increase in deviation potentially raises the frequency of the signal, thereby negating the INRs training according to spectral bias. 
Note that this effect is less significant than the effects of skewness of the distribution. Therefore, we should aim to lower the variance while maintaining a range-defined and symmetric distribution.

\begin{figure*}[!t]
    \centerline{\includegraphics[width=\textwidth]{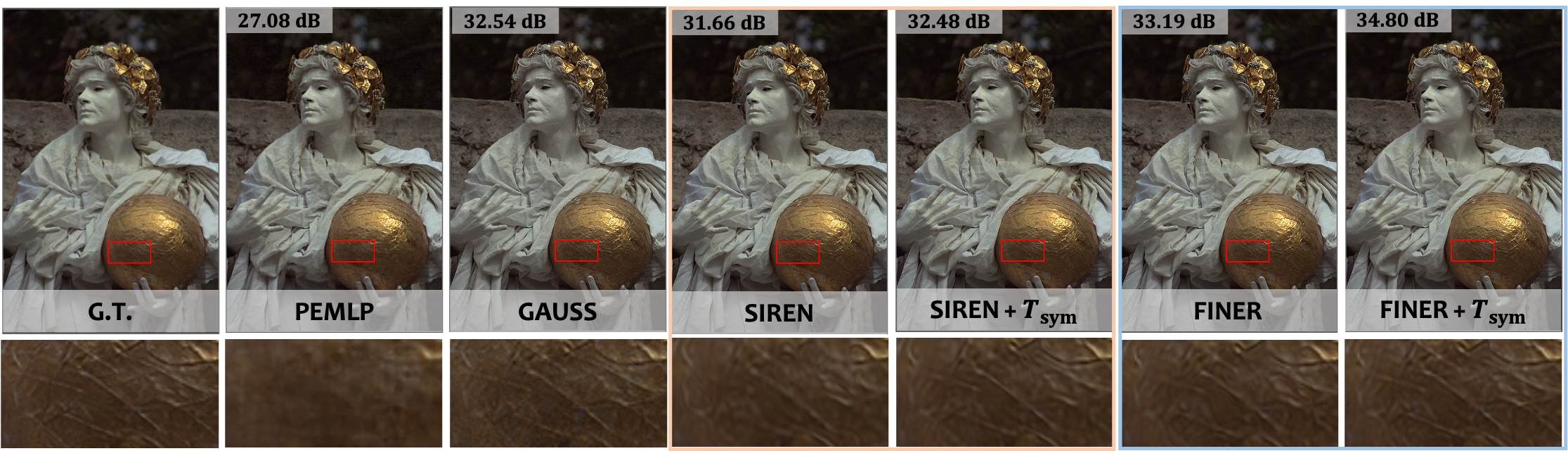}}
    \caption{Visualization for 2D natural image fitting. With \trans~in the SIREN and FINER backbones, texture details are reconstructed with greater fidelity.}
    \label{fig:image_detail}
\end{figure*}

\subsection{\Trans}

According to the Range-Defined Symmetric Hypothesis, we propose \trans~$T_\mathrm{sym}$, a nonlinear and reversible transformation to enhance the training of INRs by simultaneously meet the range-defined and symmetric criteria. We define the basic form of $T_\mathrm{sym}$ as follows:
\begin{equation}
    \label{eq:sym}
    T_\mathrm{sym}(\mathbf{y}) = (b-a)\mathbf{y}_{0}^\beta + a, \mathbf{y}_{0} = \frac{\mathbf{y} - \min({\mathbf{y})}}{\max(\mathbf{y}) - \min(\mathbf{y})}.
\end{equation}
Here $a$ and $b$ are the minimum and maximum of target bounding $I$. $\beta$ is the power of the 0-1 normalized data $\mathbf{y}_{0}$. Given the monotonic increasing nature of power functions, $T_\mathrm{sym}$ is a reversible transformation within the bounding $I=[a, b]$. We can use $\beta$ to control the symmetry of $\mathbf{y}$, which is defined $\beta$ as:
\begin{equation}
    \label{eq:beta}
    \beta = \frac{\log \lambda}{\log{(\mathrm{Q}_{\lambda})}}, \ \  
    F_Y(\mathrm{Q}_{\lambda}) = P(Y \leq \mathrm{Q}_{\lambda}) = \lambda,
\end{equation}
where $F_Y$ is the cumulative distribution function of the signal $\mathbf{y}$. Note that for the natural signal, it is impossible to completely transform $\mathbf{y}$ into a symmetric distribution and recover it without information loss. Therefore, we set $\lambda=0.5$ to approximate symmetry while maintaining reversibility and defined range. After the power transformation of $\beta$, we will have $\mathrm{E}[\mathbf{y}_0^\beta] = 0.5,$ $\mathbf{y}_0^\beta \sim [0,1]$, and thus have $\mathrm{E}[T_\mathrm{sym}(\mathbf{y})] = \frac{1}{2}(a+b)$ (symmetric) and $T_\mathrm{sym}(\mathbf{y}) \sim I=[a,b]$ (range-defined).

Directly implementing the transformation as depicted in Eq.~\ref{eq:sym} can indeed enhance implicit neural representations based on our hypothesis. However, given the complex distribution of natural signals, there are side effects in some cases. Therefore, we propose two methods to improve the adaptability and robustness of \trans, named deviation-aware calibration and adaptive soft boundary, to address the extreme boosting of deviation and boundary issues after \trans.

\textbf{Deviation-Aware Calibration.} According to the supplementary statement of our hypothesis, an increase in deviation will cause a performance drop after setting to symmetric. 
When there is an extreme increase in deviation at the boundaries, the negative impact may outweigh the benefits of \trans, resulting in decreased performance. 
To address this issue, we propose deviation-aware calibration. Given the 0-1 normalized signal $\mathbf{y}_0$, $\beta$ would be calibrated as follows:
\begin{equation}
    \label{eq:cali}
    \beta^{+} = \beta - \Delta\beta =  \beta - \xi\int_{0}^{\tau} [f_{\mathbf{sym}}(y) - f(y)] \mathrm{d}y,
\end{equation}
where $f(\cdot)$ and $f_{\mathbf{sym}}(\cdot)$ are the probability density functions for $\mathbf{y}_0$ and $\mathbf{y}_0^\beta$, respectively. $\xi$ and $\tau$ are scaling factors that determine the degree and scope of deviation to be calibrated. Note that Eq.~\ref{eq:cali} only works for $\beta > 1$ (squeeze to the left side), and there is $\Delta\beta = \xi\int_{1-\tau}^{1} [f_{\mathbf{sym}}(y) - f(y)] \mathrm{d}y$ when squeezing to the right side, corresponding to $\beta < 1$.

\textbf{Adaptive Soft Boundary.} Power transformation can maintain the same bounding for 0-1 normalized data because of its property of being monotonic increasing and its invariance when the base is 0 or 1. However, when setting rather large $\beta$, this invariance can result in a gap~(e.g., $1^{5}=1$ and $0.9^{5}=0.59$), which breaks the intrinsic continuity of the data and thus lowers the performance. 
We alleviate this problem by simply adding a soft padding between the boundaries. 
This soft boundary helps compress the range inward, thereby preventing the large gradients produced by power functions at the boundaries from disrupting continuity.
Given that this padding will somewhat break the range-defined condition, we set it to be adaptive to the percentage of the intensity of the boundary values, $f(0)$ and $f(1)$. We rewrite the 0-1 normalization in Eq.~\ref{eq:sym} as:
\begin{equation}
    \label{eq:soft}
    \mathbf{y}_0^+ = \frac{\mathbf{y} - [1+\kappa f(0)]\min(\mathbf{y})}{[1+\kappa f(1)]\max(\mathbf{y}) - [1+\kappa f(0)]\min(\mathbf{y})},
\end{equation}
where $\kappa$ is the scaling factor to control the range of the soft boundary.

\textbf{Complete Formulation.} Integrating these two methods, Eq.~\ref{eq:sym} can be rewritten as:
\begin{equation}
\label{eq:final_sym}
T_\mathrm{sym}(\mathbf{y}) = (b-a)[\mathbf{y}_0^+]^{\beta^+} + a,
\end{equation}
constituting the complete formulation of \trans.

\textbf{Analysis for Skewness Reduction.} 
Here we provide a preliminary analysis of \trans's effectiveness in reducing signal skewness and improving INR training. We simplify the analysis by modeling natural signal asymmetry using Log-normal distribution.
Consider a signal $\mathbf{y} \sim \text{Lognormal}(\mu, \sigma^2)$ where $\mu \in (0,1)$ and $\sigma >0$, with skewness $\gamma_y = (\text{exp}(\sigma^2) +2)\sqrt{\text{exp}(\sigma^2) -1}$. Let the transformed signal be $\mathbf{z} = T_{sym}(\mathbf{y})$ as defined by Eq.~\ref{eq:final_sym}.
Given the property of Log-normal distribution, we have $\beta={\log \lambda}/{\log \left(Q_\lambda\right)} \in (0,1)$. Therefore, $\ln(\mathbf{z}) = \ln(\mathbf{x}^\beta) \sim \mathcal{N}(\beta\mu, \beta^2\mu^2)$. The skewness of the transformed signal is  $\gamma_z = (\text{exp}(\beta^2\sigma^2) +2)\sqrt{\text{exp}(\beta^2\sigma^2) -1} < \gamma_y$, demonstrating a clear reduction in skewness. 
For $\beta \in (1, +\infty)$, similar proof follows by transforming the signal as $\mathbf{y}^{\prime} := \max(\boldsymbol {\zeta} - \mathbf{y},\boldsymbol{0})$, where $\boldsymbol \zeta$ can be set to $\Phi^{-1}(0.95)$ and $\Phi(\cdot)$ denotes the cumulative distribution function of $\mathbf{y}$.

%% file: sections/4_experiment.tex
\section{Experiments}
\label{sec:trans}

\begin{table*}[tbp]
    \centering
    % \small
    \begin{tabular}{l|ccc|cc|cc|cc|cc}
        \toprule
        & \multicolumn{3}{c|}{Attributes}  & \multicolumn{2}{c|}{0.5k iterations} & \multicolumn{2}{c|}{1k iterations} & \multicolumn{2}{c|}{3k iterations} & \multicolumn{2}{c}{5k iterations}\\ 
        Trans. & STD & Skew & Range &  PSNR$\uparrow$ & SSIM$\uparrow$  &  PSNR$\uparrow$ & SSIM$\uparrow$  &  PSNR$\uparrow$ & SSIM$\uparrow$ &  PSNR$\uparrow$ & SSIM$\uparrow$ \\

        \midrule
  
        0-1 norm.          & 0.24 &0.31  & 0.5 &24.46& 0.9584 &  27.95 & 0.9782 & 31.95 & 0.9898 & 33.60 & 0.9924     \\
        z-score std.       & 1.00  &0.31 & 2.2 &26.49 & 0.9716 & 29.63 & 0.9841& 33.71& 0.9924 & 35.64  & 0.9945     \\
        \midrule
        gamma~($0.5$)   & 0.41 &-0.28 &1.0 &26.14 & 0.9727 & 29.88 & 0.9879 &33.89& 0.9953 & 35.41 &0.9953    \\
        gamma~($2.0$)  & 0.48  &1.08  &1.0 & 24.34 & 0.9409& 27.45& 0.9590&30.59 & 0.9713& 31.81     &0.9745    \\
        \midrule
        scale~($\times 0.5$)  & 0.24&0.31  &0.5 & 26.01& 0.9694& 30.56& 0.9866& 34.28&  0.9930 &35.51   &0.9942    \\
        scale~($\times 1.0$)    & 0.49 &0.31  &1.0 & 26.25&0.9704 & 30.23 & 0.9860& 34.70& 0.9935 &36.18    &0.9948   \\
        scale~($\times 2.0$)   & 0.97 &0.31  &2.0  & 26.19 &0.9692 & 29.11& 0.9820&33.02 & 0.9913 &34.99  & 0.9938    \\
        \midrule
        inverse        & 0.49 &-0.31   & 1.0  &26.51 & 0.9807& 30.44 &0.9897& 34.88& 0.9946 &36.37    & 0.9955   \\
        RPP        & 0.49 &0.31  &1.0 & 13.86 & 0.8045 & 17.04& 0.9024 &20.27 & 0.9531 & 21.09     & 0.9613  \\ 
        box-cox  & 0.47 &-0.19 & 1.0  &26.00 & 0.9729 & 30.03& 0.9879& 34.15 &0.9941&35.47   & 0.9954   \\ 
        \midrule
        \rowcolor{gray!20}
        \textit{sym}-power  & 0.46 &0.16  &1.0 & 26.96 & 0.9750&31.18 & 0.9888& 35.41 & 0.9949 &36.75   & 0.9959  \\
        \bottomrule
    \end{tabular}
     \caption{
    {{Comparison of different transformations for INRs. Our method~(\textit{sym}-power) consistently achieves the best performance compared to all other transformations throughout each phase of training. }}
    }
    \label{table:trans}
\end{table*}

In this section, we verify that \trans~can freely enhance INR training, surpassing all existing data transformations in reconstruction performance.
Additionally, the effectiveness of the method is demonstrated across various modalities, including 1D audios, 2D images, and 3D videos.
We implemented all experiments using $l_2$ loss functions and the Adam optimizer~\cite{adam}.
% For different backbones, we followed the default settings from the official implementations unless otherwise stated. 
All experiments were conducted on 4 GPUs equipped with NVIDIA RTX 3090.
% , using the PyTorch framework~\cite{torch}.

\subsection{Comparison of Different Transformations} 
We reviewed the various data transformations employed in current INR training, as well as those studied in prior work. These include 0-1 normalization, z-score standardization, linear ranging, inverse operations, and random pixel permutation (RPP), among others.
We also included the Box-Cox transformation~\cite{box-cox} in our comparison, a widely-used nonlinear method to improve the normality and symmetry of the data.
We compared all of these transformations with our method under the same settings to verify its effectiveness. We chose SIREN as the backbone, fitting the widely used processed DIV2K datasets~\cite{div2k, pe} and Kodak dataset~\cite{kodak}. Following the same setting as~\cite{finer}, we used a 3 $\times$ 256 MLP for all comparisons and maintained other settings same strictly. 
We set the total number of iterations to 5000, with hyper-parameters $\xi=0.5$, $\tau=0.1$, and $\kappa=256$ in our method.
We use the above setting for all other experiments unless otherwise stated.

\noindent \textbf{Metrics.}
We used PSNR and SSIM as evaluation metrics, both evaluated at iteration 5k. Additionally, we measured the skewness and range of transformed data using the third central moment and the ratio of transformed bounding to target bounding. These metrics are represented as:
\begin{equation}
\begin{aligned}
\mathrm{Skew}(T(\mathbf{y})) &= E\left[\left(\frac{T(\mathbf{y}) - \mu}{\sigma}\right)^3\right], \\ 
\mathrm{Range}(T(\mathbf{y})) &= \frac{\max{T(\mathbf{y})} - \min{T(\mathbf{y})}}{b - a},
\end{aligned}
\end{equation}
where $I = [a, b] = [\min(\sin(\cdot)), \max(\sin(\cdot))] = [-1, 1]$. 
The effectiveness of transformations can be quantified by how closely the skewness approaches 0 and the range approaches 1, indicating alignment with our hypothesis. 
% Our analysis employs metrics averaged across all samples.

\noindent \textbf{Results.} 
The results of the comparison are demonstrated in Tab.~\ref{table:trans} and Tab.~\ref{table:trans_kodar}, respectively on natural and Kodak datasets. 
% Our method achieves superior reconstruction quality compared to existing transformation methods. 
Quantitative analysis through standard deviation (STD), skewness, and range metrics validates our method's effective approximation of desired distribution properties, explaining its state-of-the-art performance. The performance degradation observed in box-cox transformation can be attributed to its lack of deviation-aware calibration, which particularly affects certain samples.
While RPP pioneered the transformation perspective for INR training optimization, its primary contribution lies in its counter-intuitive insights rather than performance improvements. Under standard conditions, RPP underperforms our method and baseline by approximately 15 dB. Although it shows marginal benefits with excessive iterations, its standard performance suffers from disruption of pixel continuity.
Other transformations demonstrate inferior performance compared to Symmetric Power Transformation due to their limitations in addressing one or more critical aspects: skewness correction, range bounding, or preservation of signal frequency and continuity properties.

\noindent\textbf{Complexity.} Performance analysis reveals minimal computational overhead, with transformation operations requiring only 0.2 seconds compared to the typical 5-minute training duration. The memory overhead is negligible, requiring only a single float-16 parameter ($\beta$).

\begin{table}[tbp]
    \centering
    % \small
    \begin{tabular}{l|cccc}
        \toprule
         & PSNR & PSNR  &  PSNR & SSIM   \\
        Trans. & (1k)$\uparrow$  & (3k)$\uparrow$ & (5k)$\uparrow$ & (5k)$\uparrow$ \\
        \midrule
        0-1 norm. & 27.54 & 31.24 & 32.76 & 0.9944 \\ 
        z-score std. & 29.29 & 33.10 & 34.84 & 0.9962 \\ 
        scale~($\times 1.0$) & 30.06 & 33.88 & 35.21 & 0.9962  \\ 
        RPP &  16.73 & 18.57 & 19.12 & 0.9329 \\
        box-cox & 30.22 & 33.93 & 35.14 & 0.9966 \\
        \midrule
        \rowcolor{gray!20}
        \textit{sym}-power & 30.79  &34.50  &35.66 & 0.9968\\

        \bottomrule
    \end{tabular}
    \caption{
    {{Performance of \textit{sym}-power on Kodak datasets.}}% 
    }
    \label{table:trans_kodar}
\end{table}

\subsection{1D Audio Fitting}
Fitting 1D audio data can be formulated as $F_{\theta}: \mathcal{X} \in \mathbb{R}^1 \mapsto \mathcal{Y} \in \mathbb{R}^{1}$. We use LibriSpeech~\cite{libri} datasets to evaluate the effectiveness of our method.
Specifically, we selected the \texttt{test.clean} split of the dataset and cropped each audio to the first 5 seconds at a sampling rate of 16k Hz. Following the setting of Siamese SIREN~\cite{siamese}, we set $\omega$ and $\omega_0$ both to 100 in the SIREN backbone. 
We used average MSE, STOI~\cite{STOI} and SI-SNR to evaluate the reconstruction quality. 
As demonstrated in Tab.~\ref{table:audio}, our method consistently improves reconstruction quality across various phases of INR training. Notably, the improvement is more significant at the early stages of training (1k iterations).

\begin{table}[t]
    \centering
    \begin{tabular}{l|ccc}
        \toprule
         Backbones & MSE $\downarrow$  & STOI $\uparrow$  & SI-SNR $\uparrow$ \\
        \midrule
        vanilla~(1k) & 2.16e-3 & 0.559 & 1.855 \\
        \rowcolor{gray!20}
        + $T_\mathrm{sym}$~(1k) & 1.63e-3 & 0.728 & 2.764\\
        \midrule
        vanilla~(3k) & 0.57e-3 & 0.863 & 8.502 \\
        \rowcolor{gray!20}
        + $T_\mathrm{sym}$~(3k) & 0.43e-3 & 0.890 & 9.561\\
        \midrule
        vanilla~(5k) & 0.29e-3 & 0.898 & 11.077 \\
        \rowcolor{gray!20}
        + $T_\mathrm{sym}$~(5k) & 0.22e-3 & 0.916 & 12.102\\
  
        \bottomrule
    \end{tabular}
    \caption{
    {{1D audio fitting task.}}%
    }
    \label{table:audio}
\end{table}

\subsection{2D Image Fitting}
\begin{figure}[!t]
    \centerline{\includegraphics[width=\columnwidth]{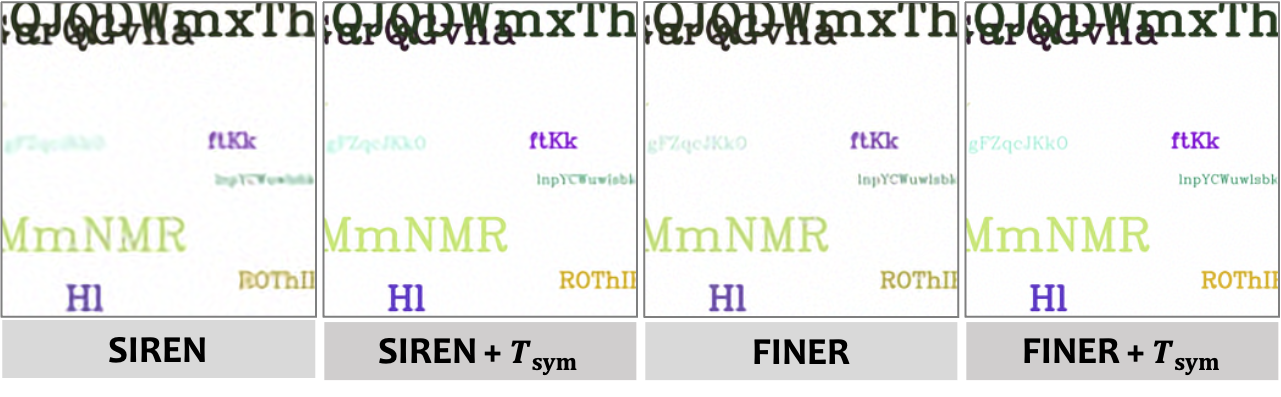}}
    \caption{Visualization for 2D text image fitting. \Trans~can enhance the clarity and color fidelity of the synthesized text.} 
    \label{fig:text_compare}
\end{figure}

Fitting 2D image data can be formulated as $F_{\theta}: (x,y) \mapsto (r,g,b)$.
We verified the effectiveness of our method in processed DIV2K, Kodak and text image dataset~\cite{pe}. We implemented \trans~on SIREN and state-of-the-art FINER to show the effectiveness of our method. We also compared our method with position encoding MLP~(PEMLP), Gauss~\cite{gauss} and WIRE~\cite{wire}. 
The qualitative results are presented in Tab.~\ref{table:kodak}~(natural image) and Tab.~\ref{table:text}~(text image). Our method improves reconstruction quality at 1k, 3k, and 5k iterations in both scenarios. 
Integrating our method with SIREN surpasses vanilla SIREN, GAUSS and WIRE in 2D text fitting. Furthermore, FINER + $T_\mathrm{sym}$ achieves state-of-the-art performance in both tasks, thanks to the improved variable-periodic functions.
Fig.~\ref{fig:image_detail} shows the visual quality comparison for 2D natural image (sculpture) fittings. With \trans~in the SIREN and FINER backbones, texture details are reconstructed with greater fidelity. 
Fig.~\ref{fig:text_compare} visualizes the results for 2D text fitting, where our method significantly enhances the clarity and color fidelity of the synthesized text.

\noindent \textbf{Why Does 2D Text Fitting Show More Significant Improvement?}
Our method achieves notably larger improvements in synthesized text reconstruction (Tab.~\ref{table:text}) compared to natural images and videos (Tab.\ref{table:kodak} and Tab.~\ref{table:video}). 
This performance difference is primarily due to our transformation's mechanism of strengthening input signal symmetry. Synthesized text data inherently has limited variety in pixel intensities; for instance, a text image with 3 words might only contain 4 pixel values (3 text colors and 1 background color). This sparse intensity distribution significantly makes the data distribution imbalance and thus lower the inherent symmetry attribute of data. Consequently, our transformation can have a more pronounced impact on synthesized text data by substantially improving symmetry. In contrast, natural images inherently maintain a degree of symmetry due to their abundant pixel intensities, resulting in relatively modest improvements when applying our method.

\begin{table}[tbp]
    \centering
    % \small
    \begin{tabular}{l|cccc}
        \toprule
         & PSNR & PSNR  &  PSNR & SSIM   \\
        Backbones         & (1k)$\uparrow$  & (3k)$\uparrow$ & (5k)$\uparrow$ & (5k)$\uparrow$ \\
        \midrule

        PEMLP     & 25.79 & 28.33 & 29.26 &  0.9787   \\                                         
        GAUSS     & 30.46 & 34.02 & 35.39 &   0.9926\\
        WIRE   &  28.86 & 32.56 & 33.91 & 0.9872\\
        \midrule
        SIREN   &  30.23 & 34.70 & 36.18 & 0.9948 \\
        \rowcolor{gray!20}
        SIREN + $T_\mathrm{sym}$ & 31.18 & 35.41 & 36.75 & 0.9959 \\
        \midrule
        FINER    & 32.30 & 36.79 & 38.44 & 0.9964\\
        \rowcolor{gray!20}
        FINER + $T_\mathrm{sym}$    &   33.04 & 37.58 & 39.03 & 0.9972  \\
        \bottomrule
    \end{tabular}
    \caption{
    {{2D natural image fitting task.}}% 
    }
    \label{table:2D_nat}
\end{table}

\begin{table}[tbp]
    \centering
    % \small
    \begin{tabular}{l|cccc}
        \toprule
         & PSNR & PSNR  &  PSNR & SSIM   \\
        Backbones & (1k)$\uparrow$  & (3k)$\uparrow$ & (5k)$\uparrow$ & (5k)$\uparrow$ \\
        \midrule

        PEMLP     & 27.36  & 29.58 & 30.45 &   0.9877   \\                                         
        GAUSS     & 31.11 &  34.40 &    35.58 & 0.9954  \\
        WIRE   &29.97 & 33.21 & 34.26    & 0.9943   \\
        \midrule
        SIREN   & 30.06 &  33.88 & 35.21    & 0.9962   \\
        \rowcolor{gray!20}
        SIREN + $T_\mathrm{sym}$ & 30.79 & 34.50 & 35.66 &  0.9969   \\
        \midrule
        FINER    & 31.99 &  36.12 &   37.49 & 0.9974\\
        \rowcolor{gray!20}
        FINER + $T_\mathrm{sym}$    & 32.83 &  36.80 & 37.95  & 0.9980    \\
        \bottomrule
    \end{tabular}
    \caption{
    {{Additional verification on Kodak datasets.}}% 
    }
    \label{table:kodak}
\end{table}

\begin{table}[t]
    \centering
    % \small
    \begin{tabular}{l|cccc}
        \toprule
        & PSNR & PSNR  &  PSNR & SSIM   \\
        Backbones & (0.1k)$\uparrow$  & (0.3k)$\uparrow$ & (0.5k)$\uparrow$ & (0.5k)$\uparrow$ \\
        \midrule

        PEMLP     &  17.71 & 19.44 & 21.11 &  0.9838    \\                                        
        GAUSS     &  23.72 & 30.48  &32.98 &  0.9962    \\
        WIRE     & 22.98 & 29.68 & 32.58     &0.9973\\
        \midrule
        SIREN   & 20.41 & 23.77 &  25.06     & 0.9926   \\
        \rowcolor{gray!20}
        SIREN + $T_\mathrm{sym}$ & 24.28 &28.89& 30.39&   0.9979  \\
        \midrule
        FINER    &  22.24 & 26.17  &  27.56  &  0.9956 \\
        \rowcolor{gray!20}
        FINER + $T_\mathrm{sym}$   & 27.64 & 32.71  & 34.30 & 0.9992    \\
  
        \bottomrule
    \end{tabular}
    \caption{
    {{2D text fitting task. }}
    }
    \label{table:text}
\end{table}

\subsection{3D Video Fitting}
Fitting 3D video data can be formulated as $F_{\theta}: (x,y,t) \mapsto (r,g,b)$. We conducted the video fitting with SIREN and FINER backbones, both using the same network size of $6 \times 256$. Following the settings of~\cite{diner}, all experiments were evaluated on the \textit{ShakeNDry} video from the UVG dataset~\cite{uvg} (the first 30 frames with 1920$\times$1080 resolution). Each scenario was trained for 100 epochs.
The results are demonstrated in Tab.~\ref{table:video}. Our method consistently improves video fitting with both the SIREN and FINER backbones in terms of PSNR and SSIM. Both metrics are averaged over all frames.

\begin{table}[t]
    \centering
    % \small
    \begin{tabular}{c|cc|cc}
        \toprule
        &  \multicolumn{2}{c|}{SIREN}& \multicolumn{2}{c}{FINER} \\ 
        Backbones & vanilla &  \cellcolor{gray!20}+ $T_\mathrm{sym}$ & vanilla & \cellcolor{gray!20}+ $T_\mathrm{sym}$ \\
        \midrule
        PSNR $\uparrow$ & 26.97 & \cellcolor{gray!20}27.25 &  27.28 & \cellcolor{gray!20}27.51\\
        SSIM $\uparrow$ & 0.9803 & \cellcolor{gray!20}0.9828 & 0.9804 & \cellcolor{gray!20}0.9831\\
        \bottomrule
    \end{tabular}
    \caption{
    {{3D video fitting task.}}% 
    }
    \label{table:video}
\end{table}

\subsection{Ablation Study}
To validate the effectiveness of each components within our proposed \trans, we conducted a series of ablation experiments. All settings are align with the experiment of comparing different transformations.
Our methods~(full $T_{sym}$) consists three primary components: basic form of \trans~(basic $T_{sym}$), deviation-aware calibration~(cali.) and adaptive soft boundary~(soft.). We conducted ablation studies for each of these components.
We used the scale~($\times 1$) as the baseline, since it is applied in the most INR framework. 
The results are presented in Tab.~\ref{table:ablation}. 
We report each studies with PSNR in iteration 1k, 3k and 5k. 
The contributions of each component can be clearly observed. 
Note that the soft boundary parts seem to contribute little to the average PSNR (0.05-0.1 dB). This is because the datasets include only a few samples that might have boundary issues, and thus their improvements are averaged out by the other data.
\begin{table}[t]
    \centering
    \begin{tabular}{l|cccc}
        \toprule
                 & PSNR & PSNR &  PSNR \\
        Settings  & (1k)$\uparrow$ & (3k)$\uparrow$ & (5k)$\uparrow$ \\
        \midrule
        baseline~(scale~$\times 1.0$)  & 30.23 & 34.70 & 36.18         \\
        \midrule    
        w/o basic $T_{sym}$  & 30.30 & 34.73 & 36.21  \\ 
        w/o cali.  & 31.07 & 35.26  & 36.61 \\
        w/o soft.  & 31.08 & 35.36 & 36.71 \\ 
        w/o cali. \& soft. & 31.04 &35.20  & 36.52  \\ 
        \midrule
        \rowcolor{gray!20}
        full $T_{sym}$  & 31.18 & 35.41 & 36.75  \\
        \bottomrule
    \end{tabular}
    \caption{
    {{Ablation experiment.}}% 
    }
    \label{table:ablation}
\end{table}

%% file: sections/2_related.tex
\section{Related Work}
\subsection{Implicit Neural Representations}
Implicit Neural Representation (INR), also known as coordinate network, can represent signals by over-fitting a neural network. Thanks to advanced progress in frequency-guided spatial encoding~\cite{pe} and effective periodic activation functions~\cite{siren}, INR has been capable of representing various data types including sketches~\cite{sketchinr}, microscopic imaging~\cite{nmi-micro}, shapes~\cite{deepsdf}, climate data~\cite{climate}, and medical data~\cite{medical-inr-survey}.
Compared with discrete representations, INR store data in a powerful and differentiable function~(MLP), which provides advantages in memory efficiency and solving inverse problems. 
This makes INR useful in various research fields, including novel view synthesis~\cite{nerf, ingp}, image enhancement~\cite{lowlight-inr}, and solving partial differential equations~\cite{pinn}.

\subsection{Enhanced Neural Field Training}
Due to spectral bias~\cite{bias} and high training cost of INR, numerous work have been proposed to improve the INR's training from various perspectives. We classify these efforts as follows.
\underline{1.Activation Functions View.} Following SIREN~\cite{siren}, which uses sine activation functions, this line of work explores different activation functions to improve INR, including Gaussian~\cite{gauss}, complex Gabor wavelets~\cite{wire}, and variable-periodic sine functions~\cite{finer}.
\underline{2.Partition-based Acceleration View.} The main idea of partition-based methods is to divide the input signal into numerous small parts and use different smaller MLPs to represent local data. The partitioning mechanisms can include regular blocks~\cite{kilonerf}, adaptive blocks~\cite{acorn}, Laplacian pyramids~\cite{miner}, and segmentation maps~\cite{partition-acc}.
\underline{3.Others.} INR can also be improved from other perspectives, including meta-learning~\cite{meta-inr}, optimizers~\cite{opt-inr}, additional modulation~\cite{incode}, reparameterized training~\cite{reparam}, etc.

\subsubsection{Data Transformation View}
Enhancing INR through data transformation is a novel perspective that explores how to transform the data itself to make it more suitable for INR, as proposed by~\cite{search}. In this work, they found that random pixel permutation (RPP) can enhance INR for high-fidelity reconstruction. Another work, Diner~\cite{diner}, adapts index rearrangement to lower the spectrum of data, thus accelerating the training process of INRs.
\textbf{Differences.} Both of these transformations are irreversible and require additional spatial cost. Our \trans~is the first method to steadily improve INR by transformaing data without incurring any additional costs.

%% file: sections/5_conclusion.tex
\section{Conclusion}
In this work, we present \trans, a novel approach to enhance INR's traning through data transformation. We introduce the Range-Defined Symmetric Hypothesis, which posits that aligning data range with activation function bounds and ensuring distribution symmetry improves INR performance. 
% This hypothesis is validated through systematic experiments with simulated data.
Guided by this hypothesis, we develop \trans, incorporating deviation-aware calibration and adaptive soft boundary mechanisms to ensure robust performance. 
% Comprehensive experiments across 1D audio, 2D image, and 3D video fitting tasks demonstrate our method's effectiveness and versatility. 
Extensive experiments demonstrate that our method achieves superior reconstruction quality over existing approaches without additional computational cost.

\noindent \textbf{Future Work.}
While our method demonstrates empirical effectiveness based on observed patterns and heuristic insights, we acknowledge the need for more rigorous mathematical foundations. Future research could establish theoretical guarantees through Neural Tangent Kernel~\cite{ntk} analysis, providing formal validation of our hypothesis and methodology.